\documentclass{article}
\usepackage{ijcai24}

\usepackage{times}
\usepackage{soul}
\usepackage{url}
\usepackage[hidelinks]{hyperref}
\usepackage[utf8]{inputenc}
\usepackage[small]{caption}
\usepackage{graphicx}
\usepackage{amsmath}
\usepackage{amsthm}
\usepackage{bbm}
\usepackage{booktabs}
\usepackage{algorithm}
\usepackage{algorithmicx}
\usepackage{algpseudocode}
\usepackage[switch]{lineno}
\usepackage{enumitem}
\usepackage{todonotes}
\usepackage[capitalise]{cleveref}
\Crefname{figure}{Fig.}{Figs.}%
\Crefname{table}{Tab.}{Tabs.}%
\Crefname{section}{Sec.}{Secs.}%
\Crefname{equation}{Eq.}{Eqs.}%
\Crefname{appendix}{Supp.}{Supps.}%
\Crefname{algorithm}{Alg.}{Algs.}%

\usepackage{amsmath,amsfonts,bm}

\def\eqref#1{equation~\ref{#1}}

\def\1{\bm{1}}

\DeclareMathAlphabet{\mathsfit}{\encodingdefault}{\sfdefault}{m}{sl}
\SetMathAlphabet{\mathsfit}{bold}{\encodingdefault}{\sfdefault}{bx}{n}

\def\sR{{\mathbb{R}}}

\newcommand{\citep}[1]{\cite{#1}}
\newcommand{\citet}[1]{\cite{#1}}

\title{Anomaly Detection of Tabular Data Using LLMs}

\author{
Aodong Li$^1$\footnote{
  The IJCAI 2024 Workshop on Anomaly Detection with Foundation Models. Correspondence to: \href{mailto:aodongl1@uci.edu}{aodongl1@uci.edu}
}
\and
Yunhan Zhao$^1$\and\\
Chen Qiu$^{2}$\and
Marius Kloft$^{3}$\and
Padhraic Smyth$^1$\and
Maja Rudolph$^2$\and
Stephan Mandt$^1$\\
\affiliations
$^1$UC Irvine\hspace{1em}
$^2$Bosch Center for AI\hspace{1em}
$^3$RPTU Kaiserslautern-Landau
}
\begin{document}

\maketitle

\begin{abstract}
Large language models (LLMs) have shown their potential in long-context understanding and mathematical reasoning. 
In this paper, we study the problem of using LLMs to detect tabular anomalies and show that pre-trained LLMs are zero-shot batch-level anomaly detectors. That is, without extra distribution-specific model fitting, they can discover hidden outliers in a batch of data, demonstrating their ability to identify low-density data regions. For LLMs that are not well aligned with anomaly detection and frequently output factual errors, we apply simple yet effective data-generating processes to simulate synthetic batch-level anomaly detection datasets and propose an end-to-end fine-tuning strategy to bring out the potential of LLMs in detecting real anomalies. Experiments on a large anomaly detection benchmark (ODDS) showcase i) GPT-4 has on-par performance with the state-of-the-art transductive learning-based anomaly detection methods and ii) the efficacy of our synthetic dataset and fine-tuning strategy in aligning LLMs to this task.

\end{abstract}

\section{Introduction}

Large language models (LLMs), which employ transformer-based architectures and billions of learnable parameters, can process and generate text that exhibits human-level realism. LLMs have enabled groundbreaking real-world applications that were hardly possible a few years ago, such as chatbots e.g.,(\href{https://openai.com/blog/chatgpt}{ChatGPT}) and code generation e.g., \href{https://github.com/features/copilot}{GitHub Copilot} \citep{roziere2023code}. 

\begin{figure}[t]
    \centering
    \includegraphics[width=0.46\textwidth]{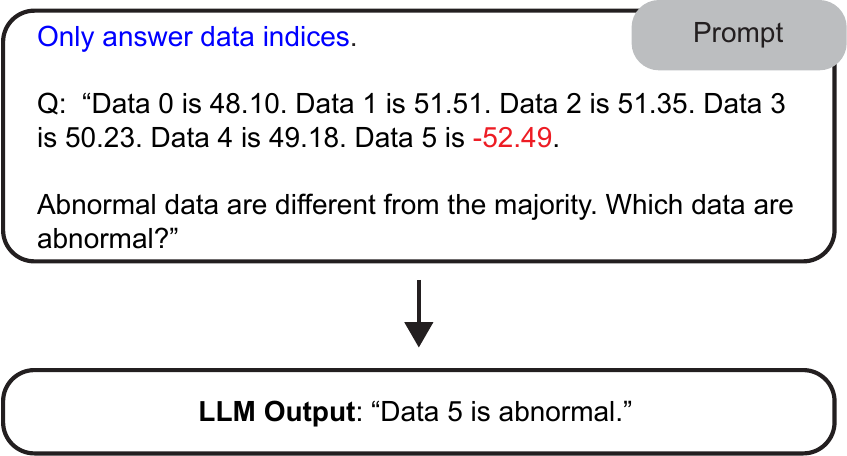}
    \caption{
    The illustration of batch-level anomaly detection with LLMs. We serialize the data batch into text and apply our proposed prompts as the input to LLMs. LLMs then respond by answering the indices of abnormal data based on LLMs' knowledge. The \textcolor{blue}{system message} ``Only answer data indices'' regularizes LLM responses and ensures responses are easy to parse. }
    \label{fig:splash}
\end{figure}

This paper studies the application of LLMs to \emph{anomaly detection} (AD)---one of the fundamental problems in machine learning occurring in many applications \citep{ruff2021unifying}. AD concerns the detection of irregular instances---so-called \emph{anomalies}---in data. There exist several settings of AD \cite{qiu2022latent,li2023deep,li2024zero}; we consider the setting of \emph{zero-shot batch-level AD} \cite{li2024zero}, illustrated in \Cref{fig:splash}, where we want to find an anomalous instance ${\bf x}_i\in\mathbb{R}^K$ in a batch of data $X=\{{\bf x}_1,\ldots,{\bf x}_N\}\subset\mathbb{R}^K$. This setting finds applications in many domains, from fraud detection and intrusion detection to medical anomaly detection and industrial damage detection.

Zero-shot batch-level AD utilizes batch information to adapt to distribution shifts and can exploit modern hardware like GPUs for parallel computation \citep{li2024zero}. Numerous shallow methods have been developed for this setting under the name of unsupervised anomaly discovery\footnote{Using ``zero-shot batch-level'' stresses that our proposed method is a deep learning-based method rather than a shallow method.} \citep{ramaswamy2000efficient,breunig2000lof,liu2008isolation,li2022ecod}. 
On the other hand, LLMs have high promise for this setting. 
Their input and output format--natural language text--leads to simpler usage for practitioners.
LLMs require no expertise in selecting anomaly detection models and setting hyperparameters.
Moreover, they have the potential to understand task background information and customize task needs. 
For example, when we know some pattern is rare but normal, we can inform LLMs to exclude that pattern from detected anomalies.

Another motivation for studying zero-shot batch-level AD arises from the data-wrangling task. \cite{narayan2022can} demonstrated employing LLMs to detect and correct errors in attribute-value pairs for tabular data, assuming that LLMs understand the attribute meanings and values as humans do. 
Unfortunately, in many real-world applications, especially in specialized domains where i) LLMs have relatively less information and ii) data are preprocessed into numerical values, LLMs cannot reliably detect errors. Therefore, we study the problem of using LLMs to detect errors in a given data batch where errors are present as outliers.

Using LLMs for zero-shot batch-level AD is challenging. First, the data consists of numerical tables, while LLMs expect text as input. Second, detecting anomalies in tables requires sophisticated computation with numerical data, such as estimating and thresholding densities. It remains unclear 1) whether LLMs can perform these tasks and 2) how to effectively prompt LLMs for AD. Third, LLMs have varying capabilities in mathematical reasoning and text understanding. 
How to align LLMs unprepared for this AD problem must be addressed.

The contributions of this paper are as follows, addressing the aforementioned challenges:
\begin{itemize}[leftmargin=*,itemsep=0pt]
    \item We propose a serialization method (illustrated in \Cref{fig:splash}) that converts batch-level anomaly detection from a numerical task to a text-based task. The method comes along without hyperparameter tuning.
    \item We empirically evaluate our approach on both synthetic and real-world data using GPT, Llama2, and Mistral. The experiments demonstrate that GPT-3.5 and GPT-4 can effectively detect anomalies in batches.
    \item We develop a strategy for fine-tuning anomaly detectors by synthesizing normal and anomalous data, thereby training the LLM to detect anomalies accurately.
    \item Experiments on the ODDS benchmark~\citep{Rayana:2016} demonstrate that our simple method using the original GPT-4 performs on par with the state-of-the-art transductive learning-based methods. The fine-tuned Mistral-based detector outperforms GPT-3.5, highlighting the effectiveness of our fine-tuning strategy. 
\end{itemize}

As follows, we discuss related works in \Cref{sec:related-word}, then present our method of applying and fine-tuning LLMs to detect anomalies in \Cref{sec:method}. We conduct experiments in \Cref{sec:exp} and conclude with \Cref{sec:conclusion}.

\section{Related Work}
\label{sec:related-word}

\paragraph{Anomaly detection with LLMs.}
\citet{gu2024anomalygpt} uses in-distribution paired images and texts to jointly train a language model and a vision encoder to describe in natural language text the found anomalies in an image.
\citep{elhafsi2023semantic} relies on LLMs' environment understanding and reasoning ability to monitor semantic anomalies in autonomous driving systems.
\citet{park2024enhancing} employs LLMs as agents to validate and interpret financial anomalies.
\citep{su2024large} surveyed the work in time series anomaly detection. 
Unlike the above work, we tackle zero-shot batch-level anomaly detection for tabular data.

\paragraph{Zero-shot batch-level anomaly detection.} 
Batch-level anomaly detection or unsupervised anomaly discovery has been studied for a long time~\cite {chandola2009anomaly}.
While numerous transductive learning-based methods have been proposed, they are shallow methods and require hyperparameter settings for each data batch~\citep{tax2004support,xu2010robust,zhou2017anomaly,ramaswamy2000efficient,li2022ecod}. 
In deep anomaly detection, zero-shot batch-level anomaly detection utilizes batch normalization layers to automatically adapt to each data batch \citep{li2024zero}.
In this work, we apply LLMs as zero-shot batch-level anomaly detectors to accomplish this task across datasets solely based on their gained knowledge through pre-training.

\paragraph{Zero-shot learning in LLMs.} LLMs have shown unprecedented zero-shot ability in many downstream NLP tasks~\citep{chang2023survey}.
Many recent works start to leverage such zero-shot ability of LLMs to other tasks, such as arithmetic reasoning~\citep{lewkowycz2022solving,imani2023industry} and time series forecasting~\citep{gruver2024large}. 
LLMs have also been applied to data wrangling for error detection~\citep{narayan2022can,vos2022towards}.
To our knowledge, we are the first to explore and benchmark LLMs on tabular anomaly detection tasks and propose effective approaches that enhance the ability of LLMs on this task.

\section{Method}
\label{sec:method}
\begin{figure*}[t]
\centering
\includegraphics[width=0.9\textwidth]{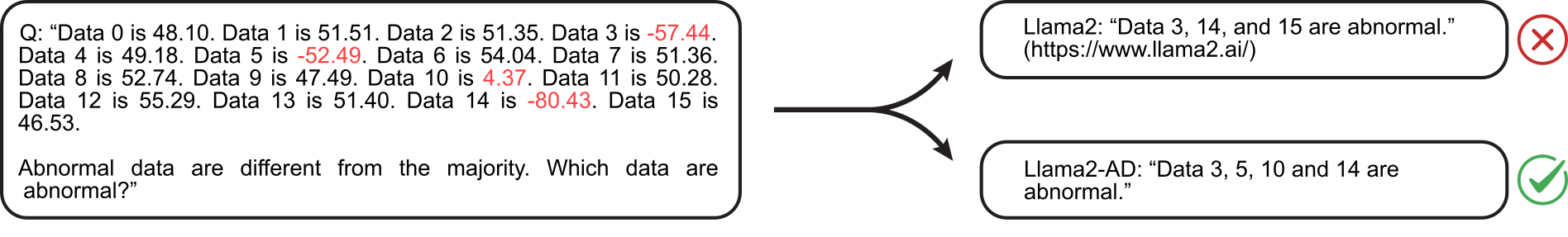}
\caption{
Illustration of Llama2 for batch-level anomaly detection before and after our fine-tuning strategy.
With the same input prompt, Llama2-70b (70-billion parameter version) makes factual mistakes--two false negatives (missing 5 and 10) and one false positive (incorrect 14). These results are obtained from \url{https://www.llama2.ai}. On the contrary, our fine-tuned 7-billion parameter (10x smaller than Llama2-70b) Llama2-AD succeeds in discovering all anomalies.
}
\label{fig:llm-pred}
\end{figure*}

This section will first present the problem setup, then introduce our text-based method for batch-level anomaly detection using large language models (LLMs), and finally propose an end-to-end fine-tuning strategy for LLMs to be better aligned to anomaly detection. 

\subsection{Problem Setup}
We consider a batch of \textit{possibly} contaminated data ${\cal D}:= \{{\bf x}_i\in\mathbb{R}^K\}_{i=1}^N$ (a numerical table) in the presence of unlabeled anomalies. 
We assume the number of anomalies is far less than normal data, i.e., the normal data takes the majority in the batch.
We stress that the data batch can contain \textit{no} anomalies. 
LLMs can tell when the batch is contaminated or not.
The aim is to identify which data points in the batch are abnormal.

\subsection{Text Formulation of Batch-level Anomaly Detection}

We assume each feature dimension is independent, and we detect anomalies for each feature separately\footnote{
  We also tried to relax this independence assumption and input the data as a vector. However, the performance degrades. The reason could be that LLMs cannot distinguish a vector from a set of scalars. For the latter, the order between elements is unimportant. 
}. The detection results of each feature dimension will then be aggregated to form the final results.

\paragraph{Data Serialization.} 
We designed a template to serialize data into text because LLMs only accept text input. 
Assuming independent features, we can detect anomalies on one feature dimension at a time. 
Then, the data to be serialized will be one-dimensional float scalars.
Denote the single-feature data by $\{x_i\in\mathbb{R}\}_{i=1}^N$. 
We use the template $T^\text{in}_i:=$"Data $i$ is $x_i$." to serialize the $i$th data point.\footnote{
  Experimental performance is not sensitive to data names.
  We also named data by ``Row'' instead of ``Data'' as if in a table where columns correspond to features or data dimensions and rows index data points.
  The experimental performance is similar.
}
The data index $i$ is necessary to disambiguate repetitive data values.
We approximate the data value up to two decimal places in the serialization.
Each serialized data point is then concatenated as input to the LLMs. 
We use ${\bf T}^\text{in}:=+\{T^\text{in}_I \}_{i=1}^N$ to denote the concatenation operation where $+$ represents concatenating each element in a set.

\paragraph{Prompt Engineering.} 
Besides the data input, we need to inform the LLMs of the anomaly detection task.
We use a description text $C:=$``Abnormal data are different from the majority. Which data are abnormal?'' to characterize anomalies and ask questions.
The serialized data input and task description together formulate the input to the LLMs, i.e., $X:=+\{{\bf T}^\text{in},C\}$.
\Cref{fig:splash} presents a serialization example with five synthetic data. 

With the input $X$, LLMs can respond to the anomaly detection request.
The response will include anomalous data indices (the numeric data indices) by design. 
In most cases, LLMs tend to generate diverse responses with long reasoning. 
We further regularize the output format by delivering another system message--``Only answer data indices.''--to the LLMs to have easy-to-parse responses.\footnote{
Use ``Only answer row numbers'' when data are named ``Row.''
}

\begin{algorithm}
\caption{LLM for batch-level anomaly detection}\label{alg:batch-ad}
\begin{algorithmic}
\Require LLM, ${\cal D}:=\{\mathbf{x}_i\in \sR^K\}_{i=1}^N$
\State Initialize anomaly score for each row $s_i = 0, i=1,\dots,N$
\For{each column $k$ in ${\cal D}$}
    \State Set serialization ${\bf T}^\text{in}$ = ``Data 1 is $\mathbf{x}_{1,k}$. Data 2 is $\mathbf{x}_{2,k}$. ... Data N is $\mathbf{x}_{N,k}$.''
    \State Set prompt $C$ = ``Abnormal data differ from the majority. Which data are abnormal?''
    \State Get response $\hat Y_k = \text{LLM}({\bf T}^\text{in}+C)$
    \State Update anomaly scores for all data points $s_i = s_i + \mathbbm{1}[i\in \hat{Y}_k]$.
\EndFor
\State \Return anomaly scores $s_i, i=1,\dots,N$
\end{algorithmic}
\end{algorithm}
\paragraph{Anomaly detection as a text-to-text task.}
One can get anomaly predictions for each feature dimension with the proposed data serialization methods and the prompts. 
We now introduce a simple method for aggregating the responses of all feature dimensions and constructing anomaly scores for each data point.

We propose to set the anomaly score of the $i$th data to be the number of occurrences of data index $i$ in all responses. 
That is, suppose the response to the $k$th feature dimension is $\hat Y_k$, then $s_i=\sum_{k=1}^K \mathbbm{1}[i\in \hat{Y}_k]$.
The anomaly scores are useful for performance evaluation and characterizing the degree of abnormality. The full procedure is presented in \Cref{alg:batch-ad}.

\paragraph{Prediction extraction from output.} 
Automatically parsing the LLM output and extracting the predicted anomalies facilitate model evaluations and improve the response-to-detection speed.
To get the predictions, we instruct the model to output only anomalous data indices by sending a system message--``Only answer data indices''
However, research shows that the capability of following instructions by LLMs differs to some extent \citep{ouyang2022training,zhou2023instruction}. 
In our experiments, we observed that the fine-tuned LLMs (e.g., Mistral-AD and Llama-AD used in the experiments)  can faithfully follow the same output format used during the fine-tuning stage. 
GPT-3.5 and GPT-4 also follow the instructions well and output succinct answers containing predicted data identifiers.
So, we can extract the predictions automatically for these models.
See \Cref{app:pred-extract-code} for script details.

The other models in our experiments, Llama-2 and Mistral, oftentimes output redundant information besides predictions even though they are instructed to only output predictions. 
Redundant information makes it hard to pinpoint the predictions without human involvement, complicating the parsing process.  
To completely suppress redundant information, we manually modify the output token probabilities at each generation step and require the generation to follow a specific pattern. 
We use regular expressions to specify the desired model output patterns with the Outlines library \citep{willard2023efficient}\footnote{
  \url{https://outlines-dev.github.io/outlines/}
}.
We found that grammar-correct formats with complete sentences are essential for generating high-quality predictions.
So the regular expression in use is \texttt{((Data [0-9]+(, [0-9]+)* are abnormal\textbackslash.)|(All data are normal\textbackslash.))} which allows the model to predict abnormal data or to abstain from predictions if all data seemingly comes from the same data-generating process.
Extracting integers from the formatted output can be accomplished by the same automatic procedure designed for GPT LLMs (see \Cref{app:pred-extract-code}).

\subsection{End-to-end Finetune Strategy}
\begin{figure}[t]
    \centering
    \includegraphics[width=0.2\textwidth]{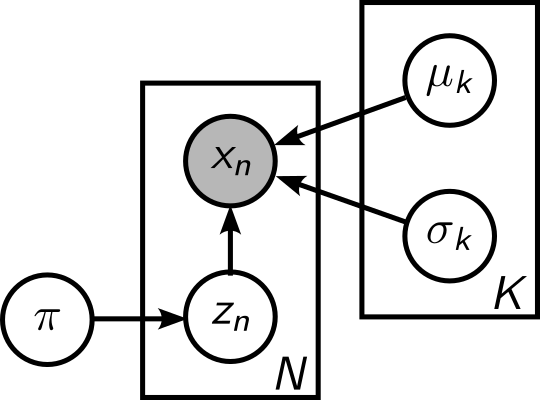}
    \hspace{0.5em}
    \includegraphics[width=0.24\textwidth]{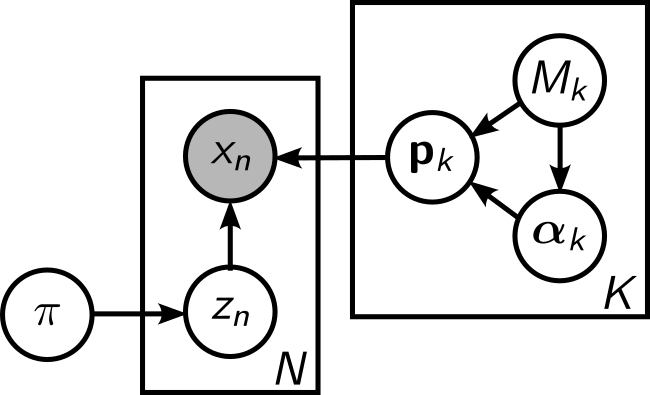}
    \caption{
    Graphical models of the synthetic data generating processes. \textbf{(Left)} We use a binary Gaussian mixture (i.e., $K=2$) to generate a batch of continuous data of size $N$. One Gaussian corresponds to normal data, and another corresponds to abnormal data. \textbf{(Right)} A multinomial mixture model ($K=2$) for discrete data where one multinomial is for normal and one for abnormal data. 
    $\pi$ controls the anomaly ratio.
    Specifics of the random variables in the models are in \Cref{app:syndata-eg}}
    \label{fig:graph-model}
\end{figure}
Unfortunately, not all LLMs are prepared to detect anomalies. \Cref{fig:llm-pred} shows a failing case with an open-sourced LLM--Llama-2 (70 billion-parameter version). Llama-2 makes factual errors: it only discovers two outliers and misses another two; it wrongly labels one normal data as abnormal. Our experiments also observed that Llama-2 may pair incorrect indices and values, generate indices beyond the batch length, or list every data as abnormal.
These phenomena signify the misalignment of Llama-2 or other LLMs in detecting anomalies.

\paragraph{Synthetic dataset.} To align LLMs in batch-level anomaly detection, we simulate a synthetic dataset with ground truth labels for LLMs to learn. The dataset contains continuous and discrete data types, covering real-world data types. 
Discrete data is a mixture of normal and abnormal Categorical distributions. Continuous data is a Gaussian mixture where normal data is a narrow Gaussian while anomalies are from a wide Gaussian. 
All the model parameters are randomly selected from a pre-defined interval.
The contamination ratio $\pi$ for both data types is also random but ensured to be smaller than 0.2. 
The data generating processes are listed in \Cref{alg:discrete-gen,alg:continuous-gen} in \Cref{app:syndata-eg}.
The corresponding graphical models are shown in \Cref{fig:graph-model}.
We simulate 2,500 batches for each data type. 
When a data batch is normal, its ground-truth response is ``All rows are normal.''\footnote{
We facilitate optimization convergence by designing high-probability response formats and using complete, grammar-consistent sentences.}
For other batches that contain anomalies, we use the ground truth answers $Y=$``Data $a_1$, $a_2$,... and $a_A$ are abnormal." where $\{a_i:y_{a_i}=1\}_{i=1}^A$ are the anomaly indices. 
Simulated synthetic data is serialized in our proposed text formulation. Synthetic data examples are in \Cref{app:syndata-eg}.

\paragraph{End-to-end fine-tune.} We align LLMs to the anomaly detection task through fine-tuning. The most common fine-tuning strategy is Chain-of-Thought~\citep{wei2022chain}. 
However, applying Chain-of-Thoughts to reason about anomalies is hard. 
Challenges arise from the complications of the AD task.
For example, suppose we construct the chain of thoughts using the two-standard deviation range method\footnote{
  The two-standard deviation range refers to the interval $[-2\sigma, 2\sigma]$ where $\sigma$ is the standard deviation. Any data points located outside this range are considered abnormal.
}. 
This method is a rough criterion and cannot cover all discrete and multimodal continuous data cases. 
In addition, asking LLMs to calculate the sample mean and sample standard deviation is another arithmetic challenge for LLMs. 

Instead, we propose to teach LLMs in an \emph{end-to-end} fashion--not focusing on ``how to solve'' but on ``what to expect.'' We directly present the answer to the model and ask LLMs to learn to predict that given answer without caring about the intermediate steps. 
Therefore, we fine-tune LLMs on the synthetic dataset $\{(X_b,Y_b)\}_{b=1}^B$ in a supervised manner. 
\Cref{fig:llm-pred} shows the efficacy of our fine-tuning method on a toy data batch. After aligning Llama2 (7 billion-parameter version) -- Llama2-AD -- detects all anomalies. 

We apply low-rank adaptation (LoRA~\citep{hu2021lora}), a parameter-efficient fine-tuning method to align LLMs. 
LoRA appends an additional low-rank weight matrix to each original weight matrix.
The low-rank matrix can be parameterized efficiently through matrix factorization.
The original weights are kept fixed during fine-tuning, and newly added low-rank matrices are updated. 
After fine-tuning, the low-rank weight matrices can be absorbed into the original weight matrix to fix the model size.

We fine-tune the LLMs by maximizing the conditional log-likelihood $ \sum_{b=1}^B\log p(Y_b|X_b;\theta_{\text{LoRA}},\theta_{\text{LLM}})$ of our simulated synthetic dataset $\{(X_b,Y_b)\}_{b=1}^B$ with respect to the learnable $\theta_{\text{LoRA}}$ while keeping LLM's original parameter $\theta_{\text{LLM}}$ fixed. 
The conditional log-likelihood can be further factorized over the tokens $\{y^b_i\}_{i=1}^{L_b}$ of each response $Y_b$ in an auto-regressive fashion: $\sum_{b=1}^B\sum_{i=1}^{L_b}\log p(y_i^b|y_{<i}^b, X_b;\theta_{\text{LoRA}},\theta_{\text{LLM}})$.
After optimization, $\theta_{\text{LoRA}}$ can be integrated into $\theta_{\text{LLM}}$ by an element-wise addition, which keeps the model size constant. More details are in \cite{hu2021lora}.

\section{Experiments}
\label{sec:exp}
\begin{table*}[t!]
        \caption{{\bf AUROC results of batch-level anomaly detection on the ODDS benchmark.} Different LLMs are evaluated. Specifically, we show the performance of two LLMs (Llama2, Mistral) before and after finetuning. Proprietary LLMs (GPT-3.5 and GPT-4) are also compared. Additional state-of-the-art transductive learning-based approaches, i.e., KNN and ECOD,  are listed for comparisons. Note that KNN and ECOD are not zero-shot batch-level methods.}
	\label{tab:odds}
        \vspace{3pt}
	\small
	\centering
	\resizebox{\linewidth}{!}{
	\begin{tabular}{lcc|cc|cc||cc}
        \toprule
        & \multicolumn{6}{c||}{Proposed Methods} &\multicolumn{2}{c}{Baselines}\\
        \cmidrule(lr){2-9}
        &GPT-3.5 &GPT-4 &Llama2 &Llama2-AD &Mistral &Mistral-AD &KNN &ECOD\\
        \midrule
abalone &78.4$\pm$15.2 &84.4$\pm$7.4 &67.2$\pm$16.9 &49.7$\pm$13.3 &73.0$\pm$15.6 &75.1$\pm$9.0 &88.0$\pm$8.8 &83.9$\pm$10.1\\
annthyroid &65.1$\pm$13.5 &82.8$\pm$4.5 &50.8$\pm$1.1 &61.5$\pm$11.9 &64.7$\pm$13.0 &82.3$\pm$9.0 &76.5$\pm$7.0 &81.4$\pm$3.3\\
arrhythmia &73.1$\pm$1.6 &75.9$\pm$3.5 &47.2$\pm$0.1 &58.7$\pm$4.9 &55.2$\pm$2.7 &61.0$\pm$4.8 &69.6$\pm$5.2 &66.3$\pm$6.7\\
breastw &63.1$\pm$34.4 &98.7$\pm$0.5 &50.4$\pm$2.4 &74.3$\pm$2.6 &62.7$\pm$4.6 &93.6$\pm$2.0 &97.5$\pm$1.0 &99.0$\pm$0.3\\
cardio &83.3$\pm$2.5 &87.1$\pm$1.4 &45.5$\pm$3.6 &71.7$\pm$10.5 &68.4$\pm$18.5 &71.7$\pm$1.5 &92.5$\pm$0.4 &95.8$\pm$1.2\\
ecoli &78.7$\pm$5.1 &73.5$\pm$2.4 &52.3$\pm$9.7 &78.9$\pm$7.3 &79.5$\pm$6.2 &79.1$\pm$4.4 &89.3$\pm$13.6 &79.1$\pm$10.1\\
forest cover &82.5$\pm$11.2 &85.9$\pm$8.1 &53.9$\pm$5.5 &58.1$\pm$25.2 &68.7$\pm$24.3 &52.4$\pm$18.8 &48.5$\pm$18.9 &83.3$\pm$3.4\\
glass &69.5$\pm$11.4 &64.2$\pm$14.1 &45.4$\pm$7.7 &56.3$\pm$4.7 &59.3$\pm$8.3 &65.9$\pm$3.9 &86.7$\pm$3.0 &68.6$\pm$8.9\\
ionosphere &83.5$\pm$2.5 &88.8$\pm$2.0 &50.7$\pm$1.4 &59.9$\pm$9.4 &64.1$\pm$2.3 &69.4$\pm$8.1 &94.7$\pm$2.5 &85.8$\pm$1.8\\
kdd &66.1$\pm$28.8 &87.4$\pm$1.6 &52.4$\pm$3.4 &58.0$\pm$3.1 &65.3$\pm$1.7 &60.1$\pm$5.6 &59.8$\pm$4.8 &88.3$\pm$1.7\\
kddrev &58.5$\pm$16.8 &72.8$\pm$5.1 &53.3$\pm$4.7 &60.7$\pm$12.6 &56.8$\pm$9.0 &50.2$\pm$14.1 &45.0$\pm$1.2 &74.4$\pm$5.3\\
letter &50.9$\pm$10.5 &53.8$\pm$1.9 &48.4$\pm$1.5 &55.3$\pm$5.6 &52.2$\pm$4.6 &50.6$\pm$2.7 &42.2$\pm$2.9 &51.0$\pm$8.3\\
lympho &90.7$\pm$5.8 &88.2$\pm$2.7 &45.1$\pm$3.8 &90.7$\pm$8.8 &74.7$\pm$9.2 &96.0$\pm$1.7 &88.4$\pm$0.0 &97.7$\pm$0.0\\
mammo &52.8$\pm$20.0 &68.7$\pm$30.3 &49.8$\pm$0.7 &55.1$\pm$13.5 &67.5$\pm$10.8 &79.9$\pm$15.4 &86.2$\pm$5.8 &94.7$\pm$5.0\\
mnist &69.9$\pm$12.0 &68.2$\pm$13.3 &48.8$\pm$1.6 &51.5$\pm$8.8 &54.6$\pm$6.3 &54.2$\pm$7.6 &54.4$\pm$3.3 &59.6$\pm$6.3\\
mulcross &86.9$\pm$8.8 &88.6$\pm$5.8 &51.4$\pm$2.0 &59.0$\pm$9.8 &60.9$\pm$7.2 &75.3$\pm$8.8 &11.1$\pm$15.7 &95.4$\pm$1.2\\
musk &75.8$\pm$9.0 &63.3$\pm$4.5 &54.3$\pm$4.0 &62.7$\pm$8.3 &65.5$\pm$14.1 &63.1$\pm$16.0 &94.9$\pm$1.0 &60.8$\pm$8.8\\
optdigits &39.5$\pm$2.6 &35.6$\pm$19.5 &58.5$\pm$12.7 &41.1$\pm$9.0 &55.8$\pm$5.7 &39.5$\pm$9.6 &24.5$\pm$15.5 &29.9$\pm$14.0\\
pendigits &49.6$\pm$3.8 &78.2$\pm$11.8 &57.2$\pm$8.6 &52.5$\pm$5.7 &56.3$\pm$11.5 &72.1$\pm$24.2 &63.4$\pm$7.1 &76.9$\pm$5.4\\
pima &55.9$\pm$6.5 &59.6$\pm$2.4 &46.0$\pm$0.8 &51.6$\pm$1.5 &55.0$\pm$4.1 &61.4$\pm$1.4 &70.9$\pm$3.5 &60.2$\pm$4.4\\
satellite &58.4$\pm$8.8 &62.7$\pm$4.9 &51.0$\pm$0.9 &58.0$\pm$6.9 &58.4$\pm$8.6 &68.3$\pm$2.4 &71.1$\pm$2.1 &60.9$\pm$6.1\\
satimage &90.5$\pm$9.1 &86.0$\pm$13.4 &53.1$\pm$7.0 &70.8$\pm$9.2 &71.7$\pm$8.0 &97.1$\pm$2.1 &94.0$\pm$7.5 &80.3$\pm$19.1\\
seismic &67.9$\pm$2.4 &68.2$\pm$3.3 &53.6$\pm$3.7 &58.4$\pm$18.3 &57.9$\pm$6.8 &70.1$\pm$4.0 &70.5$\pm$2.9 &69.2$\pm$6.2\\
shuttle &94.1$\pm$6.2 &98.9$\pm$1.1 &50.1$\pm$0.3 &72.2$\pm$7.2 &75.6$\pm$11.7 &97.5$\pm$2.2 &95.1$\pm$6.8 &98.6$\pm$1.7\\
speech &51.2$\pm$23.4 &44.9$\pm$34.3 &55.0$\pm$9.3 &37.9$\pm$19.9 &40.5$\pm$18.4 &47.8$\pm$7.7 &54.7$\pm$31.1 &61.7$\pm$25.1\\
thyroid &88.8$\pm$9.7 &95.2$\pm$2.9 &42.5$\pm$9.8 &84.5$\pm$9.4 &81.3$\pm$11.7 &92.5$\pm$2.2 &98.7$\pm$1.3 &98.3$\pm$1.1\\
vertebral &57.9$\pm$3.0 &51.6$\pm$11.4 &48.8$\pm$3.9 &48.2$\pm$4.5 &54.1$\pm$5.7 &45.3$\pm$2.6 &34.6$\pm$2.3 &44.0$\pm$3.3\\
vowels &40.9$\pm$8.1 &65.9$\pm$3.1 &51.9$\pm$6.3 &51.4$\pm$19.1 &47.3$\pm$3.5 &52.5$\pm$5.2 &96.1$\pm$3.9 &62.6$\pm$14.3\\
wbc &79.2$\pm$5.6 &93.4$\pm$2.2 &48.2$\pm$4.8 &61.3$\pm$5.4 &68.5$\pm$7.3 &88.6$\pm$8.2 &93.7$\pm$2.0 &91.9$\pm$2.3\\
wine &47.6$\pm$11.5 &51.3$\pm$10.2 &50.6$\pm$9.3 &51.2$\pm$3.9 &55.5$\pm$8.4 &59.7$\pm$12.0 &30.0$\pm$0.0 &64.9$\pm$0.0\\
\midrule
average
     &68.3$\pm$1.2 &74.1$\pm$2.2 &51.1$\pm$2.5 &60.0$\pm$3.2 &62.4$\pm$2.7 &69.1$\pm$1.0 &70.7$\pm$0.9 &75.5$\pm$1.0\\
        \bottomrule
	\end{tabular}
	}
\vspace{-0.5em}
\end{table*}

\begin{figure}[t]
\centering
\includegraphics[width=0.45\textwidth]{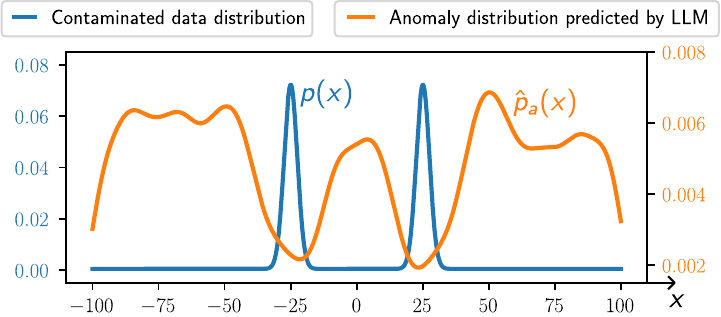}
\caption{LLMs can detect low-density regions in a contaminated data distribution. We use our Mistral-AD fine-tuned based on Mistral as the demonstrating LLM. Normal data distribution is represented by two Gaussian distributions located at -25 and 25 respectively. The contaminated data distribution is formed by combining the normal distributions and a wide uniform distribution spanned over interval $[-100, 100]$, where the contamination ratio is 0.1, resulting in $p(x)$ in blue. We sample 500 independent batches from $p(x)$ and ask the LLM to predict anomalies using our proposed method for each batch. We collect all the predicted anomalies and estimate the density by a kernel density estimator, shown by $\hat{p}_a(x)$ in orange. $\hat{p}_a(x)$ successfully captures three low-density regions of $p(x)$, demonstrating the LLM's ability to detect anomalies. More details are in \Cref{app:exp-detail}.}
\label{fig:low-density}
\end{figure}
This section shows experimental results on the ODDS anomaly detection benchmark. 
One surprising result is that our simple prompt engineering method with the original GPT-4 performs similarly to the state-of-the-art anomaly detection method. 
Our alignment method using synthetic data on Llama2 and Mistral demonstrates significant improvements over their primitive counterparts.

We first introduce the global experimental setups and then the implementation details of our proposed methods. 
Finally, we present the results.

\subsection{Experiment Setup}

We follow the widely adopted ODDS tabular data benchmark~\citep{Rayana:2016} to evaluate LLMs batch-level anomaly detection performance. 
Some LLMs have input token limits due to the context window size and GPU memory constraint. Therefore, we randomly sub-sample 150 rows and use the first 10 columns for each dataset to perform the evaluation.
We extensively study various LLMs to support our findings. Specifically, we evaluate the popular GPT-3.5 and GPT-4.\footnote{
  Specifically, we use api of gpt-3.5-turbo-1106 and gpt-4-1106-preview.
} 
We also compare two open-source LLMs, Llama2~\citep{touvron2023llama} and Mistral~\citep{jiang2023mistral}, using the 7B parameter version available at HuggingFace. We also include the LLMs (Llama2-AD and Mistral-AD) fine-tuned using our synthetic dataset and fine-tuning strategy. Lastly, we include two transductive learning approaches, KNN~\citep{ramaswamy2000efficient} and ECOD~\citep{li2022ecod}, to demonstrate better how LLMs-based methods stand against state-of-the-art approaches. 
See \Cref{app:exp-detail} for implementation details.

\subsection{Implementation Details} 
We run all experiments three times with different random seeds. 
All our experiments except GPT-3.5 and GPT-4 are performed using an A6000 GPU with PyTorch. Llama-2 and Mistral can fit into the GPU memory. The temperature and $top\_p$ generation hyperparameters are set as 0.75 and 0.9 for Llama-2 and Mistral, respectively. On the other hand, for GPT-3.5 and GPT-4, we use their default hyperparameter settings and perform the experiments through their API.

We fine-tune Llama-2-7B and Mistral-7B using LoRA parameter-efficient fine-tuning strategy~\citep{hu2021lora} on our synthetic datasets.
We generate training and validation sets separately. 
The training set involves 5000 data batches (2500 continuous data batches and 2500 discrete data batches), while the validation set contains 400 data batches (200 for continuous and 200 for discrete data). 
We finetune Llama-2-7B for five epochs and Mistral-7B for two epochs with the same learning rate 1e-3. All optimizations are convergent on the validation set. 
The resulting models are named Llama2-AD and Mistral-AD.

\subsection{Results}
\paragraph{Qualitative results.} 
Accomplishing the anomaly detection task requires LLMs to identify the low-density data of ${\cal D}$. 
To illustrate LLM's low-density region detection ability, we simulate a synthetic data distribution contaminated by anomalies. We use a two-component Gaussian mixture as the normal data distribution. We contaminate this normal data distribution with a wide uniform distribution representing abnormal data distribution. 
The final distribution is shown by $p(x)$ in blue in  \Cref{fig:low-density}. 
We sample data batches from this contaminated data distribution and
apply our fine-tuned Mistral-AD (see below) to predict anomalies.
We collect the predicted anomalies from all batches and use the kernel density estimator to fit a density $\hat{p}_a(x)$ on them.
\Cref{fig:low-density} shows $\hat{p}_a(x)$ captures the three low-density regions in $p(x)$, separated by two peak Gaussian distributions, demonstrating LLM's low-density region detection ability. 

\paragraph{Quantitative results.}
The results of OODS benchmark are shown in \Cref{tab:odds}. The results summarize two salient conclusions: (i) {\it Sophisticated LLMs are state-of-the-art zero-shot batch-level anomaly detectors.} 
Comparing GPT-4 against ECOD, state-of-the-art method on ODDS benchmark, GPT-4 shows on-par performance without extra fine-tuning, indicating the huge potential of LLMs in the anomaly detection task.
(ii) {\it Proposed end-to-end fine-tuning strategy significantly boost the performance.} 
Checking the performance of the same LLM before and after fine-tuning (Llama2 vs. Llama2-AD, Mistral vs. Mistral-AD), both models show significant improvements: on average, 8.9 and 6.7 AUROC increases, respectively, showing the efficacy of our fine-tuning strategy.

\section{Conclusion}
\label{sec:conclusion}
We consider using large language models (LLMs) to detect anomalies for numerical data wrangling. We address this problem through batch-level anomaly detection. We developed a text formulation for LLMs to accomplish this task. We found LLMs are capable to identify low-density regions in a batch of data. Surprisingly, GPT-4 is a strong zero-shot batch-level anomaly detectors that have comparable performance with state-of-the-art transductive learning methods. For LLMs that are not well aligned to this task, we designed and simulated a synthetic dataset to fine-tune the LLMs in an ``end-to-end'' fashion. Experiments demonstrate the significance of our findings and the efficacy of our proposed fine-tune strategy.

\bibliographystyle{named}
\bibliography{refs}

\appendix

\section{Synthetic Dataset}
\label{app:syndata-eg}

\subsection{Data Generating Processes}
\begin{algorithm}[th]
    \centering
    \caption{Generate discrete data }
    \label{alg:discrete-gen}
    \begin{algorithmic}[1]
        \Require hyperparameters $N^l, N^h, \pi^l, \pi^h, M^l, M^h, \alpha$
        \State $N\sim{\cal U}\{N^l:N^h\}$
        \State $\pi\sim{\cal U}_{[\pi^l, \pi^h]}$
        \State $M_n,M_a\sim{\cal U}\{M^l : M^h\}$
        \State $p_n\sim\text{Dir}(\{\alpha\}_{M_n})$
        \State $p_a\sim\text{Dir}(\{\alpha\}_{M_a})$
        \For{$i=1,\dots,N$}
        \State $x_i\sim \{(1-\pi)p_n, \pi p_a\}$
        \EndFor
        \State \Return $\{x_i\}, i=1,\dots,N$
    \end{algorithmic}
\end{algorithm}
\begin{algorithm}[th]
    \centering
    \caption{Generate continuous data}
    \label{alg:continuous-gen}
    \begin{algorithmic}[1]
        \Require hyperparameters $N^l, N^h, \pi^l, \pi^h, \mu^l, \mu^h, \sigma_n^l, \sigma_n^h$
        \State $N\sim{\cal U}\{N^l:N^h\}$
        \State $\pi\sim{\cal U}_{[\pi^l, \pi^h]}$
        \State $\mu_n, \mu_a\sim{\cal U}_{[\mu^l, \mu^h]}$
        \State $\sigma_n\sim{\cal U}_{[\sigma_n^l, \sigma_n^h]}$
        \State $\sigma_a=10\sigma_n$
        \For{$i=1,\dots,N$}
        \State $x_i\sim (1-\pi){\cal N}(\mu_n, \sigma_n^2) + \pi{\cal N}(\mu_a, \sigma_a^2)$
        \EndFor
        \State \Return $\{x_i\}, i=1,\dots,N$
    \end{algorithmic}
\end{algorithm}

Data generating processes of synthetic discrete and continuous data are presented in \Cref{alg:discrete-gen} and \Cref{alg:continuous-gen}, respectively. Discrete data is a mixture of normal and abnormal Categorical distributions. Continuous data is the clutter setup where normal data is sampled from a narrow Gaussian distribution while anomalies are from another wide Gaussian distribution. 
In practice, we generate the discrete data by setting the hyperparameters $N^l=20, N^h=100, \pi^l=0.01, \pi^h=0.2, M^l=1, M^h=4, \alpha=20$. For continuous data generation, we choose $N^l=20, N^h=100, \pi^l=0.01, \pi^h=0.2, \mu^l=-100, \mu^h=100, \sigma_n^l=0.5, \sigma_n^h=5$.
For both data types, the contamination ratio $\pi$ is smaller than 0.2.

\subsection{Data Examples}
\begin{figure*}[th]

\small
\includegraphics[width=0.98\textwidth]{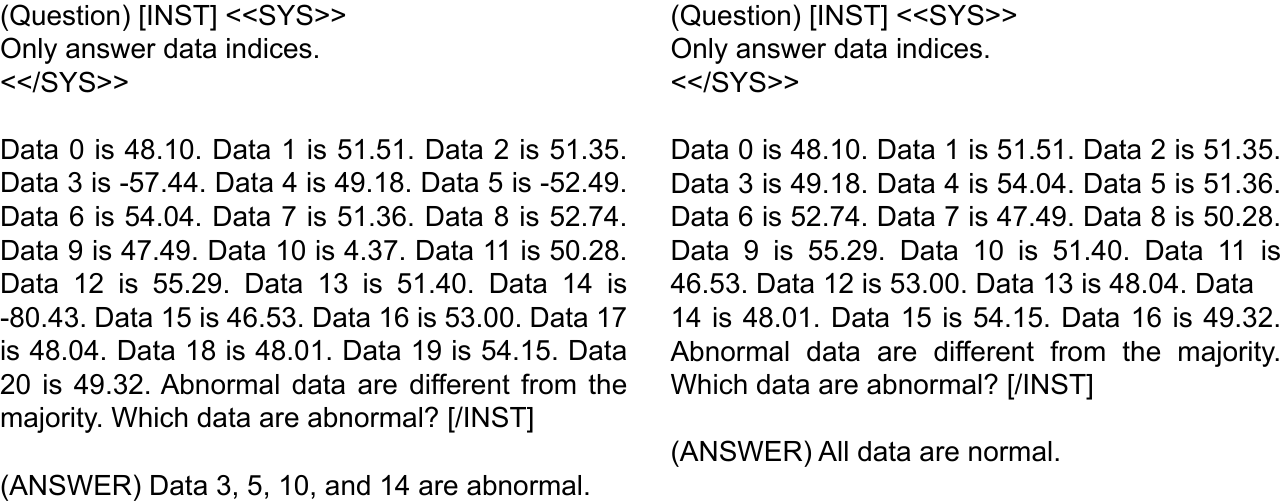}
\caption{Examples of the synthetic data for fine-tuning.}
\label{fig:syn-data-examples}
\end{figure*}

~\Cref{fig:syn-data-examples} demonstrates the prompt (Llama2 and Mistral format) and output from LLMs with generated synthetic data.

\section{Implementation Details}
\label{app:exp-detail}

\subsection{Prediction Extraction Procedure}
\label{app:pred-extract-code}
The automatic procedure for extracting model predictions in all experiments is the following code snippet in Python. 
\begin{verbatim}
    def parse_generation_results(ans, max_num=149):
        response_ret = []
        if ans.endswith("."):
            ans = ans.rstrip(".")
        ans = ans.rsplit(":->", 1)[-1]
        if ":" in and:
            ans = ans.replace(":", " ")
        ans = ans.replace(",", "")
        ans = ans.split()
    
        if "no" in ans or "No" in ans or "None" in and:
            return []
    
        for r in and:
            if r.isnumeric() and "." not in r and int(r)<= max_num:
                response_ret.append(int(r))
                
        return response_ret
\end{verbatim}

\subsection{Qualitative Study}
In \Cref{fig:low-density}, we use $p(x)=0.45{\cal N}(-25, 2.5^2)+0.45{\cal N}(25, 2.5^2)+0.1\text{Unif}(-100,100)$. $\hat{p}_a(x)$ is estimated by a kernel density estimator with 5.0-bandwidth Gaussian kernels. The predicted anomalies are collected from 500 independent batch predictions, where each batch contains 50 data points sampled from $p(x)$.

\subsection{Quantitative Study}
{\bf Implementation details.} 
The output from LLMs are naturally diverse and less controllable.
A system prompt: ``Only answer row numbers.'' is passed to all LLMs to easier parse the responses for evaluation. 
We manually filter unreasonable predictions of LLMs. Specifically, (i) we ignore predictions that beyond provided data samples; (ii) we choose the semantic consistent one if the output contains multiple answers.
We repeat all experiments 3 times with different random seeds. 
All our experiments are implemented with PyTorch using A6000 GPU. For Llama-2 and Mistral, the temperature and $top\_p$ generation hyperparameters are set as 0.75 and 0.9, respectively. For GPT-3.5 and GPT-4, we use the default hypereparameter settings.

We fine-tune all models using LoRA parameter-efficient fine-tuning strategy~\citep{hu2021lora}.
We finetune Llama-2 for five epochs and Mistral for two epochs with the same learning rate 1e-3. All optimizations are convergent.

\end{document}